\pdfoutput=1

\documentclass[11pt]{article}

\usepackage[]{acl}

\usepackage{times}
\usepackage{latexsym}

\usepackage[T1]{fontenc}

\usepackage[utf8]{inputenc}

\usepackage{microtype}

\usepackage{graphicx}
\usepackage{xspace}
\usepackage{todonotes}
\usepackage{inconsolata}
\usepackage{xcolor}
\usepackage{subfig}
\usepackage{float}

\usepackage{listings}
\usepackage{xcolor}

\colorlet{punct}{red!60!black}
\definecolor{background}{HTML}{FAFAFA}
\definecolor{delim}{RGB}{20,105,176}
\colorlet{numb}{magenta!60!black}

\lstdefinelanguage{json}{
    basicstyle=\normalfont\ttfamily,
    numberstyle=\scriptsize,
    stepnumber=1,
    numbersep=8pt,
    showstringspaces=false,
    breaklines=true,
    frame=lines,
    backgroundcolor=\color{background},
    literate=
     *{0}{{{\color{numb}0}}}{1}
      {1}{{{\color{numb}1}}}{1}
      {2}{{{\color{numb}2}}}{1}
      {3}{{{\color{numb}3}}}{1}
      {4}{{{\color{numb}4}}}{1}
      {5}{{{\color{numb}5}}}{1}
      {6}{{{\color{numb}6}}}{1}
      {7}{{{\color{numb}7}}}{1}
      {8}{{{\color{numb}8}}}{1}
      {9}{{{\color{numb}9}}}{1}
      {:}{{{\color{punct}{:}}}}{1}
      {,}{{{\color{punct}{,}}}}{1}
      {\{}{{{\color{delim}{\{}}}}{1}
      {\}}{{{\color{delim}{\}}}}}{1}
      {[}{{{\color{delim}{[}}}}{1}
      {]}{{{\color{delim}{]}}}}{1},
}

\definecolor{purple}{HTML}{B238DF}
\definecolor{green}{HTML}{14A07E}

\newcommand{\square}{{\textsc{SQuARE}}\xspace}
\newcommand{\squaretitle}{{\textsc{UKP-SQuARE}}\xspace}

\title{\squaretitle v2\\Explainability and Adversarial Attacks for Trustworthy QA}

\author{
\begin{minipage}[t]{\textwidth}
\centering
\normalsize
\bf
Rachneet Sachdeva\thanks{{ } 
Equal Contribution.
}$^{\,\,\,}$,
Haritz Puerto$^{*}$,
Tim Baumgärtner,
Sewin Tariverdian,
Hao Zhang,
Kexin Wang,
Hossain Shaikh Saadi,
Leonardo F. R. Ribeiro,
Iryna Gurevych \\
{\footnotesize \normalfont 
Ubiquitous Knowledge Processing Lab  (UKP Lab), \\Department of Computer Science and Hessian Center for AI (hessian.AI), \\Technical University of Darmstadt \\
\url{www.ukp.tu-darmstadt.de}
} 
\end{minipage}
}

\begin{document}
\maketitle
\begin{abstract}
Question Answering (QA) systems are increasingly deployed in applications where they support real-world decisions. However, state-of-the-art models rely on deep neural networks, which are difficult to interpret by humans. Inherently interpretable models or post hoc explainability methods can help users to comprehend how a model arrives at its prediction and, if successful, increase their trust in the system. Furthermore, researchers can leverage these insights to develop new methods that are more accurate and less biased. In this paper,  we introduce \square v2, the new version of \square, to provide an explainability infrastructure for comparing models based on methods such as saliency maps and graph-based explanations. While saliency maps are useful to inspect the importance of each input token for the model's prediction, graph-based explanations from external Knowledge Graphs enable the users to verify the reasoning behind the model prediction. In addition, we provide multiple adversarial attacks to compare the robustness of QA models. With these explainability methods and adversarial attacks, we aim to ease the research on trustworthy QA models. \square is available at \url{https://square.ukp-lab.de}.\footnote{The code is available at \url{https://github.com/UKP-SQuARE/square-core}}
\end{abstract}

\section{Introduction}
The recent explosion of Question Answering datasets and models is pushing the boundaries of QA systems and making them widely used by the general public in virtual assistants or chatbots \citep{rogersqa21}. This ubiquitous adoption is making regulators start preparing policies for artificial intelligence with special emphasis on explainability and robustness to adversarial attacks.\footnote{\url{https://digital-strategy.ec.europa.eu/en/policies/european-approach-artificial-intelligence}}

There are multiple methods to explain the predictions of AI models \citep{danilevsky-etal-2020-survey} and analyze their robustness \citep{zhang2020adversarial}. Some explainability methods focus on specific input attributions such as attention- and gradient-based saliency maps \citep{DBLP:journals/corr/SimonyanVZ13}. Others design interpretable models instead of using post hoc methods \citep{yasunaga-etal-2021-qa}. Lastly, most approaches that analyze the robustness of AI systems are based on \textit{adversarial attacks}, i.e., the use of inputs such as questions with minor modifications that change the system's output.

\begin{figure}[t]
\centering
\includegraphics[width=\linewidth]{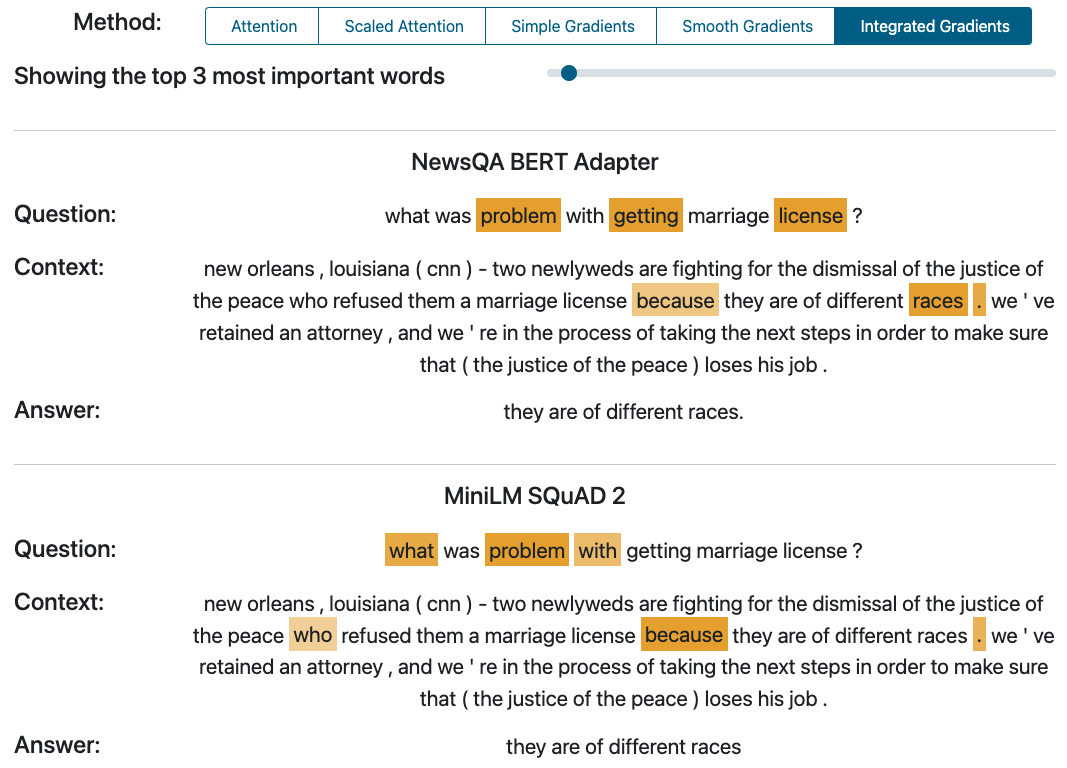}
\caption{Visualization of two saliency maps computed using integrated gradients. The darker the highlighting color, the higher its importance to get the prediction. Hovering on a word shows its importance value.}
\label{fig:saliency_map}
\end{figure}

However, exploring and comparing these methods is not straightforward for most models. Researchers usually need to manipulate libraries and create interfaces to compare them in a satisfactory manner, which is a slow and complicated process that hinders the research in trustworthy QA. 

The \square platform \citep{baumgartner-etal-2022-ukp} simplifies the process of comparing QA models by empowering NLP researchers with an online platform to deploy, run, and compare the most common QA pipelines while removing technical barriers such as model and infrastructure configurations. It includes dozens of models of multiple types, namely open-domain, extractive, multiple choice, and abstractive QA. However, the only explainability method currently implemented is behavioral test (Appendix \ref{appendix:square}), limiting the comparison between QA models based solely on the models' final predictions.




In this work, we propose \square v2, a new online platform for trustworthy QA research implementing various explainability, interpretability, and robustness methods and interfaces to facilitate research in trustworthy QA models. Specifically, we make the following contributions: 1) \square v2 supports the comparison of models based on different post hoc explainability methods. We create interactive saliency maps that illustrate the importance of each input token for the model's prediction \citep{DBLP:journals/corr/SimonyanVZ13}. 2) We extend the Datastores to include support for knowledge graphs (KG), deploy QA-GNN \citep{yasunaga-etal-2021-qa}, an interpretable graph-based model, and create an interactive visualization graph. 3) \square v2 further provides various adversarial attacks, which change the prediction by modifying the input but keeping its semantics in order to evaluate the robustness of QA models \citep{ebrahimi-etal-2018-hotflip}.



\section{Related Work}
AllenNLP demo\footnote{\url{https://demo.allennlp.org}} is the closest system to \square v2. They provide a web interface to interact with their library, where users can explore explainability functionalities \citep{gardner-etal-2018-allennlp, wallace-etal-2019-allennlp}. However, only two non-Transformer models include saliency maps and attack methods. In addition, users cannot deploy their models on this web demo, and instead, they would need to install their library and create their own interface.

Among the explainability libraries, Captum \citep{kokhlikyan2020captum} is of special relevance. It is a model interpretability library for PyTorch that includes multiple saliency maps and provides built-in visualizations. However, it does not provide a user interface to run all their methods and compare them at a glance. On the other hand, it provides an adversarial attack method, Fast Gradient Sign Method \citep{FGSM}, and some variants; however, these are not designed for NLP. 


Lastly, there are some efforts to ease the study of adversarial attacks on NLP models. Textattack \citep{morris-etal-2020-textattack} is a library that supports several attacks and is model agnostic. However, they do not provide a web interface, so users must therefore create their own visualizations in order to be able to easily compare attacks on multiple models.

In summary, \square is a single entry-point for NLP practitioners to analyze, compare, and teach QA through models' outputs, explainability, and robustness with a user-friendly interface.

\section{UKP-SQuARE}
\label{appendix:square}
\square \citep{baumgartner-etal-2022-ukp} is an open-source, online platform for NLP researchers to share, run, compare, and analyze their QA models. The platform implements a flexible and scaleable microservice architecture containing four high-level services:
\begin{enumerate}
    \item \textbf{Datastores:} Provides efficient access to large-scale background knowledge such as Wikipedia.
    
    \item \textbf{Models:} Allows the dynamic deployment and inference of a wide variety of models implemented in the Hugging Face transformers library \citep{wolf-etal-2020-transformers} or adapters \citep{pfeiffer-etal-2020-adapterhub}.
    
    \item \textbf{Skills:} Implements a configurable QA pipeline (e.g. multiple-choice, open-domain, or extractive QA) leveraging the Datastores and Models service. They can be added dynamically by the users to the system.
    \item \textbf{Explainability:} provides a set of unit tests (questions and answers in our case) \citep{ribeiro-etal-2020-beyond} to compare the predictions with the expected answers and, in this way, analyze the biases and weaknesses of the Skills. 
\end{enumerate}
\square is designed to ease the comparison and analysis of models. Users can deploy their models using a simple interface without the need of any code and then, they can compare outputs of different models side-by-side. This paper describes a new major update of SQuARE.

\section{Trustworthy Methods for QA}

Modern neural networks have significantly improved in performance in recent years; however, their explainability have not followed the same improvement \citep{rogers-etal-2020-primer}. Additionally, despite their impressive performance, the models are vulnerable to adversarial attacks. The goal of \square v2 is to provide the research community with a set of tools to facilitate the research on trustworthy QA. \square simplifies and provides visualizations for saliency maps, graph-based interpretable models, and adversarial attacks. The following sections briefly describe the methods provided in \square.

\begin{figure}[t]
\centering
\includegraphics[width=\linewidth]{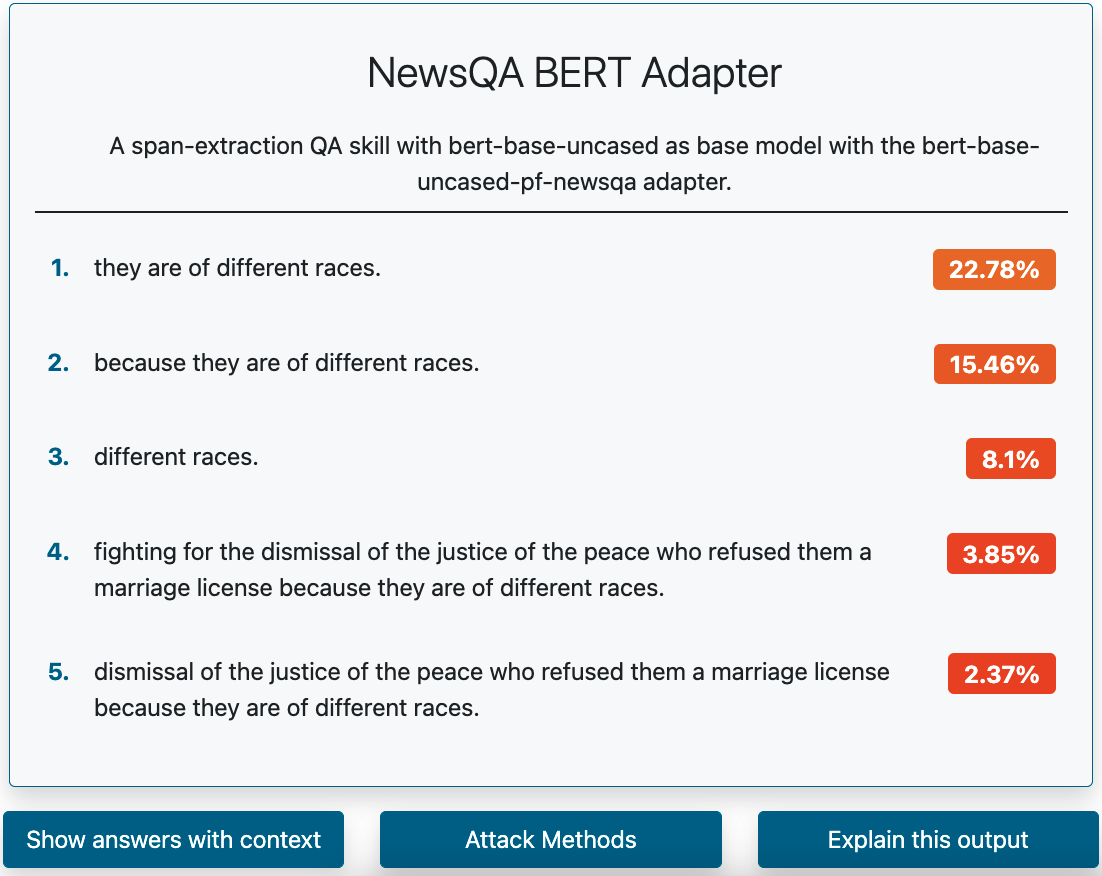}
\caption{The "Explain this output" and "Attack Methods" are shown under the predictions of the Skill.}
\label{fig:explain_btn}
\end{figure}

\subsection{Saliency Maps}
\label{sec:saliency_maps}
Saliency Maps assign an attribution weight to the input tokens to assess their importance in the model prediction, as illustrated in Fig.~\ref{fig:saliency_map}. To obtain this visualization, a user needs to click on the button \textit{"Explain this output"} located after the predictions of any Skill, as shown in Fig.~\ref{fig:explain_btn}.

In \square, we use two families of attribution methods to construct saliency maps: i) Gradient-based methods and ii) Attention-based methods.

\subsubsection{Gradient-based Methods}\label{sec:gradient-based-methods}
A common approach to obtaining an importance score for the input tokens is to compute the gradients on the embedding layer against the model prediction. The magnitude of the gradient corresponds to the change of the prediction when updating the embedding. Therefore, a large gradient has a large effect on the prediction, indicating the importance of the input.

\textbf{Vanilla Gradient} \citep{DBLP:journals/corr/SimonyanVZ13} utilizes the plain gradients of the embedding layer of the model as importance weights of the inputs.

\textbf{Integrated Gradient} \citep{10.5555/3305890.3306024} integrates the straight line path from the vector of zeros to the input token embedding. The value of this integral is the weight of this token to make the prediction since it represents the amount of information given with respect to the zero vector (i.e., no information).

\textbf{SmoothGrad} \citep{DBLP:journals/corr/SmilkovTKVW17} adds gaussian noise to the input to create multiple versions and then average their saliency scores. In this way, this method can smooth the saliency scores and alleviate noise from local variations in the partial derivatives.

\subsubsection{Attention Methods}
\label{sec:attention_methods}
Neural NLP models have broadly incorporated attention mechanisms, which are frequently recognized for enhancing transparency and increasing performance \citep{vaswani2017attention}. These methods compute a distribution over the input tokens that can be considered to reflect what the model believes to be important. Following \citep{jain-etal-2020-learning}, we build a saliency map using the average \textbf{attention weights} of the heads from the CLS token to the other tokens of the input. However, \citet{serrano-smith-2019-attention} argue that attention weights are inconsistent and may not always correlate with the human notion of importance. Thus, they propose an alternative, \textbf{Scaled Attention}, which we also integrate in \square, that multiplies the attention weights by their corresponding gradients to make it more stable.

\subsection{Interpretable Graph-based Models}
\label{sec:graph_viz}
\begin{figure}[t]
\centering
\includegraphics[width=\columnwidth]{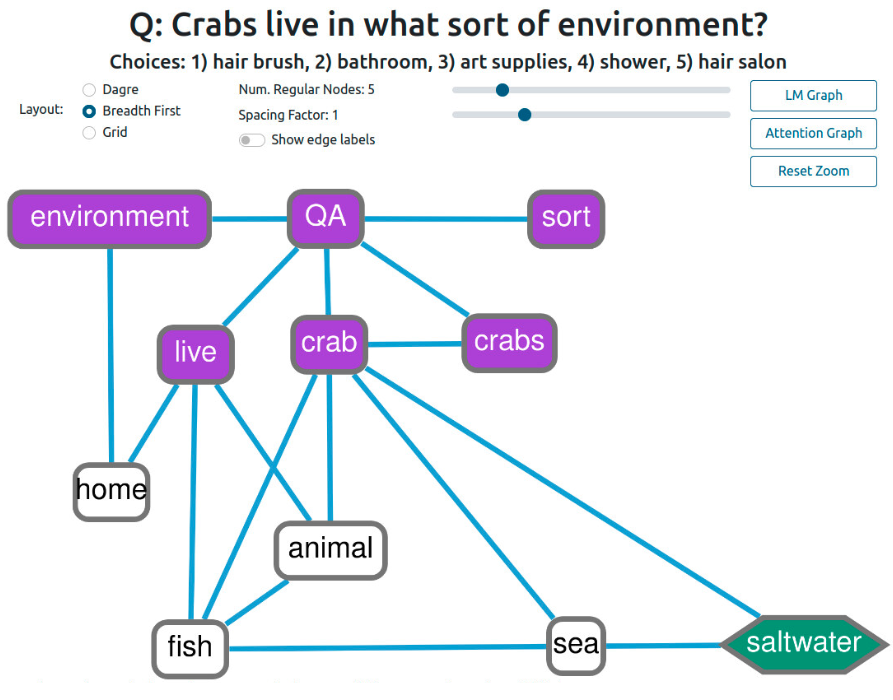}
\caption{Visualization of the graph used by QA-GNN Skill to answer the question. Question nodes in \textcolor{purple}{purple}, answer nodes in \textcolor{green}{green}.}
\label{fig:graph_viz}
\end{figure}

Knowledge graphs store knowledge in the form of relations (edges) between entities (nodes). In addition to the explicit facts they represent, they enable explainable predictions by providing reasoning paths \citep{yasunaga-etal-2021-qa}. In \square v2, we deploy QA-GNN \citep{yasunaga-etal-2021-qa}, a graph-based QA model, as a Skill (more details on Appendix~\ref{appendix:qa-gnn}) and ConceptNet \citep{Speer_Chin_Havasi_2017} as a graph Datastore. Since QA-GNN uses a KG (i.e., ConceptNet) for QA reasoning, it is possible to analyze its working graph to identify the most important entities and relations for the answer prediction. As shown in Fig.~\ref{fig:graph_viz} and later discussed in \S\ref{sec:use_cases-graph}, we provide an interface that enables the visualization of the graph-based reasoning process executed by the model. 

\paragraph{User Interface.}
In order to plot the graphs, \square provides users with a \textit{"Show graph"} button after the predictions of the QA-GNN Skill at the bottom of the page. Clicking on this button displays a modal window with multiple options to render the graph, as shown in Fig.~\ref{fig:graph_viz}. Controls include a switch to show or hide edge labels, a slider to show the top $k$ nodes, another slider to select the spacing factor between nodes, and a group of radio buttons to select the layout (Dagre\footnote{\url{https://github.com/cytoscape/cytoscape.js-dagre}}, Breath First, and Grid). In addition, we offer two types of visualizations: i) a graph where the nodes are sorted by the relevance scores generated by the model and ii) a graph with the nodes sorted by the sum of the attention scores of their incoming edges.

\subsection{Adversarial Attacks}\label{sec:adversarial_attacks}
Adversarial attacks make use of inputs that expose vulnerabilities of machine learning models to understand their robustness and identify how to improve them \citep{ebrahimi-etal-2018-hotflip}. To simplify the exploration of adversarial attacks on a wide range of Skills, we implement the following four methods in \square for span-extraction Skills and leave the other Skills for future updates.

\begin{figure}[h]
\centering
\includegraphics[width=\linewidth]{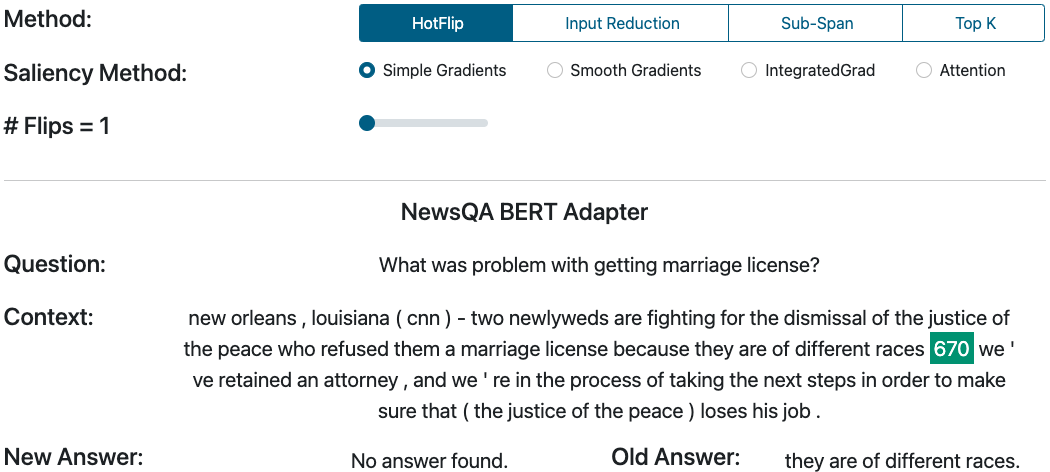}
\caption{HotFlip Attack. Changing one word changes the prediction.}
\label{fig:hotflip}
\end{figure}

\textbf{HotFlip} \cite{ebrahimi-etal-2018-hotflip} uses a saliency method (\S\ref{sec:saliency_maps}) to score input words and subsequently replaces the top words with semantically similar words to alter the prediction of the model. An example of the interface is shown in Fig.~\ref{fig:hotflip}. The words highlighted in green are replacements, and when hovering over them, the original word is shown in a tooltip.

\textbf{Input Reduction} \cite{feng-etal-2018-pathologies} iteratively removes unimportant words from the question based on their saliency scores (\S\ref{sec:saliency_maps}), without changing the model's prediction. An example is shown in Fig.~\ref{fig:inputReduction} (Appendix~\ref{appendix:advesarial_attack_figures}).

\textbf{Sub-Span} \citet{jain-etal-2020-learning} computes the saliency scores of the input words to select a contiguous span that maximizes the accumulative saliency score and uses this span as an explainability method. Instead, we leverage this method to create an adversarial attack. We identify a sub-span of the context that explains the output and use it as the whole context. In this way, the model has the key information, such as a phrase containing the answer, but not the whole context. Therefore, it is possible to identify a sub-span that lacks the nuance to answer the question properly, but since the answer occurs in the sub-span, the model may retrieve it due to spurious correlations. Fig.~\ref{fig:subspan} (Appendix~\ref{appendix:advesarial_attack_figures}) and \S~\ref{sec:use_cases-attacks} show an use case of this attack.

\textbf{Top K}. Similarly as in the previous case, \citet{jain-etal-2020-learning} compute the saliency scores of the input words to identify the top $k$ words from the context that explains the output answer. We leverage this method to create an adversarial attack. While the top $k$ words are key to obtaining the answer, they are usually not contiguous. Therefore, creating a new context by concatenating these words yields a grammatically and semantically incorrect text. If the model still identifies the correct answer using this new context, it would be due to spurious correlations. An example of this attack is shown in Fig.~\ref{fig:topk} (Appendix~\ref{appendix:advesarial_attack_figures}).

\paragraph{User Interface}
After the user queries any Skill, the button \textit{"Attack Method"} is shown under the predictions, as shown in Fig.~\ref{fig:explain_btn}. After clicking on it, a modal page is shown where users can conduct adversarial attacks.

\section{Datastores for Knowledge Graphs}
\label{sec:datastores}
To best re-use the existing Datastore while being efficient and robust, we rely on an Elasticsearch instance to store KGs. In particular, we represent nodes and edges as documents and include information to recreate the graph structure, such as their connectivity. We show the schema of these documents in Appendix \ref{sec:appendix-KG-scheme}.

In addition, we implement two main functionalities: Firstly, users can dynamically add and update new KGs as long as the structure of the KG can be converted to the schema shown in Appendix \ref{sec:appendix-KG-scheme}. This allows supporting any KG that is requested by the community. For demonstration purposes, we provide ConceptNet \citep{Speer_Chin_Havasi_2017} as a built-in KG. More information is available on the Datastores documentation\footnote{\url{https://square.ukp-lab.de/docs/api/datastores/}}. Secondly, we implement a subgraph extraction method. Given a list of root nodes (e.g., the entities in a question), it extracts all the nodes and edges in the vicinity of $k$ hops to the roots. Since ConceptNet is densely connected, we limit the maximum number of hops to 3. However, this is a parameter that can be adjusted for any KG. Lastly, after the extraction, we prune the disconnected nodes.

\section{Case Study}
\subsection{Saliency Maps}
Our new saliency map interface allows users to compare the explanation of the outputs of up to three Skills. As shown in Fig.~\ref{fig:saliency_map}, thanks to this visualization, we can easily observe that the first Skill, \textit{NewsQA BERT Adapter}, gives the correct answer for the right reasons since it identifies \textit{"races"} as a keyword. Even though the second Skill, \textit{MiniLM SQuAD 2}, also returns the correct answer, the Skill does not seem to understand the context properly. In particular, the most important words for the predictions are not related to the answer. We argue that this interface can provide insights into whether the model understands the task and thus make the Skills more trustworthy.

\subsection{Interpretable Models}
\label{sec:use_cases-graph}
ConceptNet provides background knowledge that can boost the commonsense abilities of NLP models. As shown in Fig.~\ref{fig:graph_viz}, the QA-GNN Skill makes use of the KG to connect the entities \textit{crab} with \textit{sea} and with \textit{saltwater}, the answer. This explicit path helps to identify why the model returns its answers. However, it still requires some human effort to interpret the graph. For example, ConceptNet does not include the triple (\textit{sea}, \textit{is a}, \textit{environment}), which could be seen as counter-intuitive. 

On the other hand, other non-graph-based Skills need post hoc explainability methods such as saliency maps (\S\ref{sec:saliency_maps}) to explain their output. However, post hoc methods have raised concerns about the possibility of not being faithful to the actual computations performed by the model or giving incomplete explanations as in saliency maps \citep{liu2021going}. In particular, saliency maps identify what parts of the input are relevant for the prediction, but they do not explain how or why the model obtains the output.

\subsection{Adversarial Attacks}
\label{sec:use_cases-attacks}
Using the Sub-span attack method shown in Fig.~\ref{fig:subspan} (Appendix~\ref{appendix:advesarial_attack_figures}), we can observe that the Skill gives the correct answer even though it does not have information about \textit{Super Bowl 50}, which is needed. A robust Skill should instead return "not enough information." This example suggests that the Skill is conducting a superficial question-context overlap matching without understanding the nuances of the question, a phenomenon previously identified by \citet{lim2020analysis}. Similarly, the input reduction attack shown in Fig.~\ref{fig:inputReduction} (Appendix~\ref{appendix:advesarial_attack_figures}) shows the same phenomenon. After removing most words from the question, the resulting question is not semantically complete, yet the Skill gives the correct answer.

\section{Conclusion and Future Work}
We present \square v2, a web platform that unifies three families of methods for analyzing QA models: saliency maps, adversarial attacks, and interpretable models. Firstly, we offer an interactive interface that allows users to compare multiple saliency map methods for all the Skills deployed in \square. Secondly, we provide an interface to conduct adversarial attacks. This interface allows the community to study the robustness of QA models. Lastly, we deploy an interpretable graph-based model and provide an interface to visualize the reasoning paths that the model may conduct. To deploy this Skill, we extend the Datastores module to support both text documents and KGs. These contributions give \square a set of tools to compare, analyze, and explain the behavior of QA models. Since \square allows the deployment of almost any Transformer-based model effortlessly, our new explainability interface empowers the community with tools for trustworthy QA research. \square is actively under development. Future updates will include new KGs such as WikiData \citep{wikidata}, automated Skill selection \citep{geigle:2021:arxiv}, and Skill collaboration \citep{puerto2021metaqa}.

\section*{Limitations}
Although saliency maps attempt to explain the output of the models, they should be analyzed with skepticism. As discussed in \S\ref{sec:attention_methods}, attention-based saliency maps may not correlate with the human interpretation of importance, and in general, they do not explain how and why the model creates the outputs. Instead, saliency maps only aim to identify regions of the input that upon removal, changes the output. 

Currently, we only deploy one graph-based model (QA-GNN) and one knowledge graph (ConceptNet). However, our Datastores \S\ref{sec:datastores} and graph visualization interface \S\ref{sec:graph_viz} are flexible enough to accommodate any other model, and thus, we invite the community to create pull requests and deploy their graph-based models on \square.

\section*{Ethics and Broader Impact Statement}
\paragraph{Intended Use.}
The intended use of \square is to facilitate the comparison of QA models through multiple angles such as performance, explainability, interpretability, and robustness. Our platform allows NLP practitioners to share their models with the community removing technical barriers such as configuration and infrastructure so that any person can reuse these models. This has a straightforward benefit for the research community (i.e., reproducible research and analysis of prior works) but also to the general public because \square allows them to run state-of-the-art models without requiring any special hardware and hiding complex settings such as virtual environments and package management.

\paragraph{Potential Misuse.}
Our platform makes use of models uploaded by the community. However, this current version does not incorporate any mechanism to ensure that these models are fair and without bias. We hope that the new tools we provide in this work can help the community understand the outputs of QA models and identify potential biases or unfair behaviors. Thus, we currently delegate the fairness checks to the authors of the models. We are not held responsible for errors, false or offensive content generated by the models. Users should use them at their discretion.

\paragraph{Environmental Impact.}
Since \square empowers the community to run publicly available Skills on the cloud, it has the potential to reduce CO\textsubscript{2} emissions from retraining previous models to make the comparisons needed when developing new models.

\section*{Acknowledgements}
We thank Irina Bigoulaeva and Haishuo Fang for their insightful comments on a previous draft of this paper. We also thank the anonymous reviewers for their insightful feedback. 

This work has been funded by the German Research Foundation (DFG) as part of the UKP-SQuARE project (grant GU 798/29-1), the QASciInf project (GU 798/18-3), and by the German Federal Ministry of Education and Research and the Hessian Ministry of Higher Education, Research, Science and the Arts (HMWK) within their joint support of the National Research Center for Applied Cybersecurity ATHENE.

\bibliography{anthology,custom}
\bibliographystyle{acl_natbib}

\clearpage
\appendix

\section{Knowledge Graph Document Schema}
\label{sec:appendix-KG-scheme}
The nodes of a knowledge graph are stored in the Datastore as a json document using the following schema:
\begin{lstlisting}[language=json,firstnumber=1]
{
    "node_id": {
        "_id": "keyword",
        "name": "keyword",
        "description": "text",
        "type": "keyword"
    }
}
\end{lstlisting}
The edges of a knowledge graph are stored in the Datastore as a json document using the following schema:

\begin{lstlisting}[language=json,firstnumber=1]
{
    "edge_id": {
        "_id": "keyword",
        "name": "keyword",
        "description": "text",
        "type": "keyword",
        "in_id": "keyword",
        "out_id": "keyword",
        "weight": "double"
    }
}
\end{lstlisting}

\section{QA-GNN Implementation} 
\label{appendix:qa-gnn}

We implement the QA-GNN inference pipeline on \square based on the official implementation of QA-GNN model.\footnote{\url{https://github.com/michiyasunaga/qagnn}} We disregard the training code since training QA models is not in the scope of \square and connect the model with the Datastores service holding the KG. This makes it more flexible for future updates of ConceptNet. Lastly, the retrieved nodes with corresponding attention weights and relevance scores are accessible along with the answer prediction. With this information, we plot the graph using the JavaScript library Cytoscape.js \citep{Franz2016CytoscapejsAG}.

\clearpage
\onecolumn

\section{Adversarial Attack Figures}\label{appendix:advesarial_attack_figures}
\newcommand{\widthAttack}{1\linewidth}


\begin{figure*}[h]
\centering
\includegraphics[width=\widthAttack]{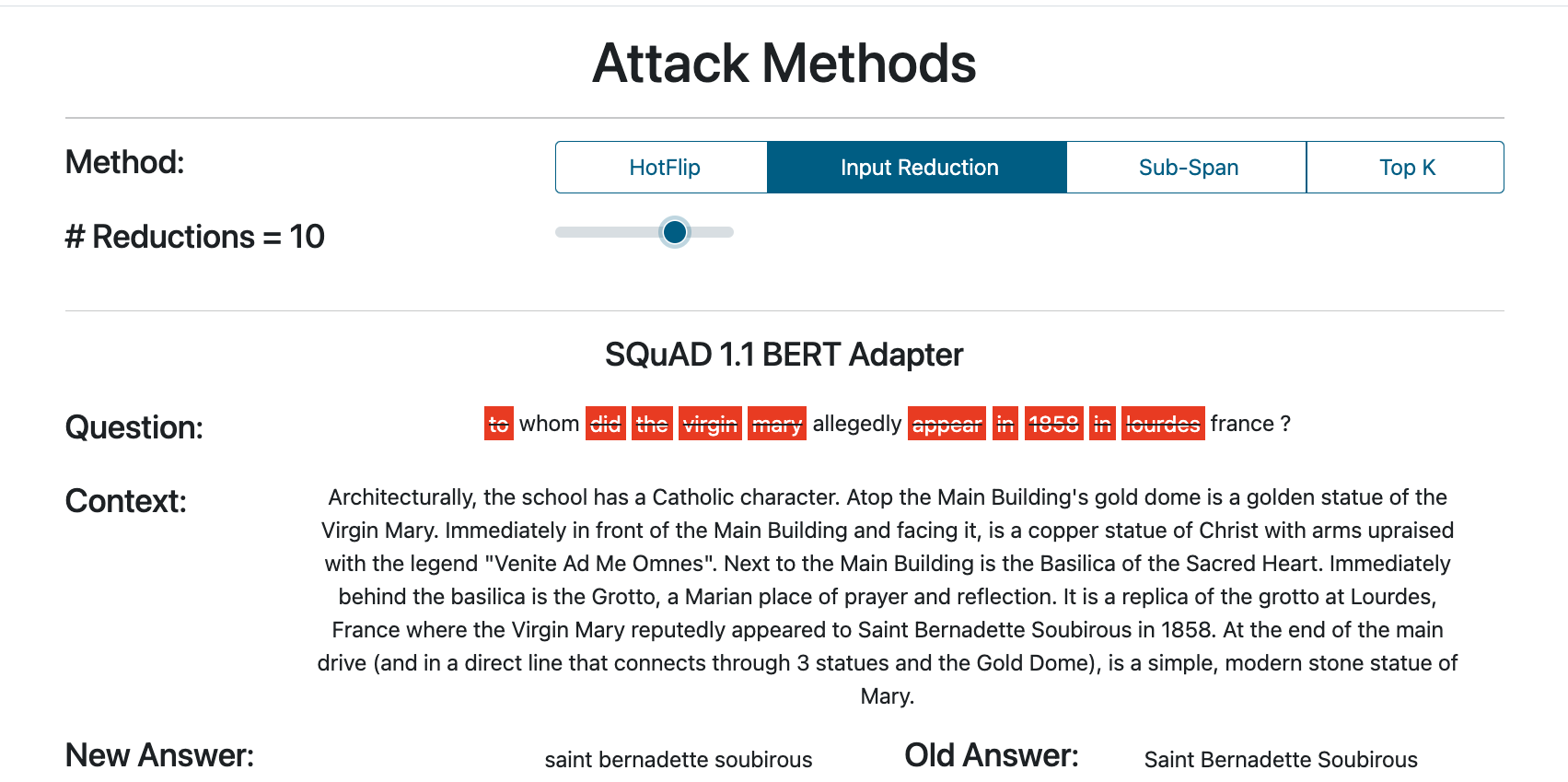}
\caption{Input Reduction. After removing tokens from the question, the new question is not specific enough to be answerable. Yet, the model still gives the same answer evidencing a spurious correlation.}
\label{fig:inputReduction}
\end{figure*}
\begin{figure*}[h]
\centering
\includegraphics[width=\widthAttack]{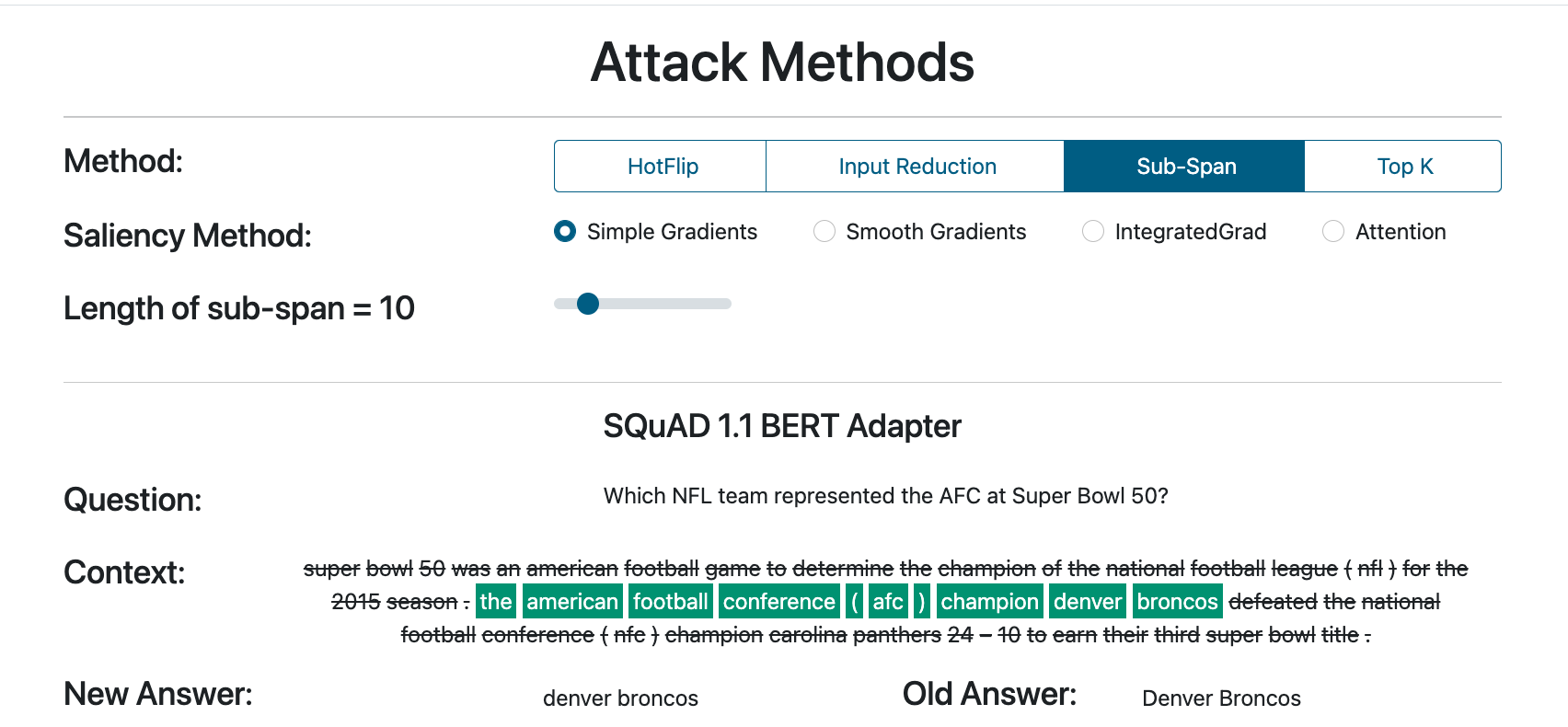}
\caption{Sub-Span Attack. Removing part of the context leaves a new context without the nuances needed to properly respond to the question (i.e., \textit{at Super Bowl 50}).}
\label{fig:subspan}
\end{figure*}
\begin{figure*}[h]
\centering
\includegraphics[width=\widthAttack]{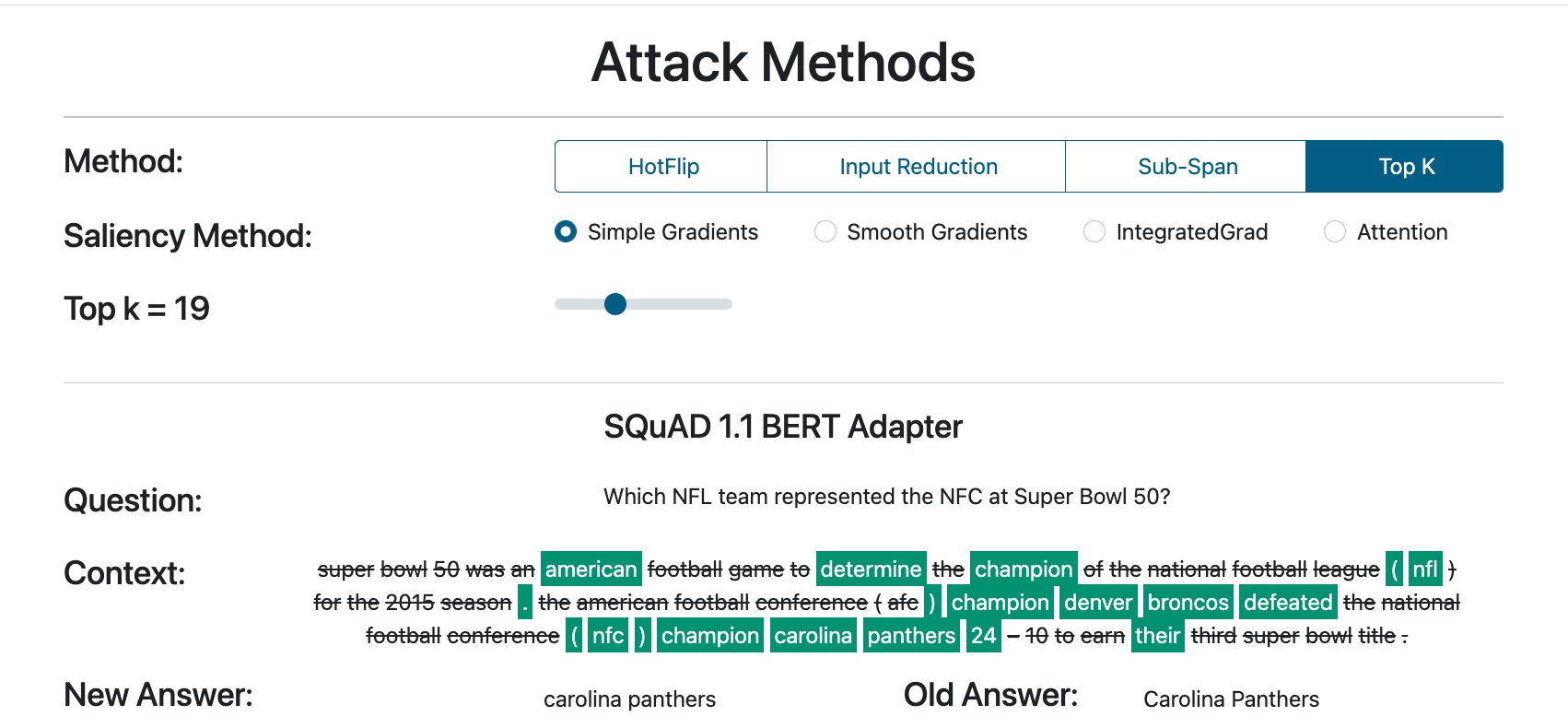}
\caption{Top $K$ Attack. Using as context the highlighted words, the Skill still gives the same answer even though the context is semantically and grammatically incomplete and does not include \textit{Super Bowl 50}.}
\label{fig:topk}
\end{figure*}

\end{document}